\documentclass[runningheads]{llncs}

\usepackage{eccv}

\usepackage{eccvabbrv}        
\usepackage{graphicx}
\usepackage{booktabs}
\usepackage{amsmath,amssymb}
\usepackage[table]{xcolor}

\usepackage{multirow}
\usepackage{multicol}
\usepackage{wrapfig}
\usepackage{url}              
\usepackage{comment}          
\usepackage{bbding}
\usepackage{enumitem}

\usepackage[breaklinks,colorlinks,citecolor=eccvblue]{hyperref}

\input{custom_env}

\newcommand{\methodname}{Multi-Cluster Memory}

\begin{document}

\title{One Pool Is Not Enough: Multi-Cluster Memory for Test-Time Adaptation}
\titlerunning{Multi-Cluster Memory for Test-Time Adaptation}

\author{%
Yu-Wen Tseng\inst{1}\textsuperscript{*}\and
Xingyi Zheng\inst{1}\textsuperscript{*}\and
Ya-Chen Wu\inst{1}\and
I-Bin Liao\inst{3}\and
Yung-Hui Li\inst{3}\and
Hong-Han Shuai\inst{2}\and
Wen-Huang Cheng\inst{1}
}

\authorrunning{Y.-W. Tseng et al.}

\institute{%
National Taiwan University, Taiwan \\
\email{\{d12922018, wenhuang\}@csie.ntu.edu.tw}
\and
National Yang Ming Chiao Tung University, Taiwan
\and
Hon Hai Research Institute, Taiwan\\
\textsuperscript{*}Equal contribution
}

\maketitle

\begin{abstract}
Test-time adaptation (TTA) adapts pre-trained models to distribution
shifts at inference using only unlabeled test data. Under the
Practical TTA (PTTA) setting, where test streams are temporally
correlated and non-i.i.d., memory has become an indispensable
component for stable adaptation, yet existing methods universally
store samples in a single unstructured pool. We show that this
\emph{single-cluster} design is fundamentally mismatched to PTTA:
a stream clusterability analysis reveals that test streams are
inherently multi-modal, with the optimal number of mixture
components consistently far exceeding one. To close this structural
gap, we propose \textbf{Multi-Cluster Memory (MCM)}, a plug-and-play
framework that organizes stored samples into multiple clusters using
lightweight pixel-level statistical descriptors. MCM introduces three
complementary mechanisms: descriptor-based cluster assignment to
capture distinct distributional modes, Adjacent Cluster Consolidation
(ACC) to bound memory usage by merging the most similar temporally
adjacent clusters, and Uniform Cluster Retrieval (UCR) to ensure
balanced supervision across all modes during adaptation. Integrated
with three contemporary TTA methods on CIFAR-10-C, CIFAR-100-C,
ImageNet-C, and DomainNet, MCM achieves consistent improvements
across all 12 configurations, with gains up to 5.00\% on ImageNet-C
and 12.13\% on DomainNet. Notably, these gains scale with
distributional complexity: larger label spaces with greater
multi-modality benefit most from multi-cluster organization.
GMM-based memory diagnostics further confirm that MCM maintains
near-optimal distributional balance, entropy, and mode coverage, whereas single-cluster memory exhibits
persistent imbalance and progressive mode loss. These results
establish memory organization as a key design axis for practical test-time adaptation.
\end{abstract}

\section{Introduction}
\label{sec:introduction}

Deep learning models trained on large-scale datasets achieve
remarkable performance within their intended scenarios, yet remain
vulnerable to distribution shifts during real-world
deployment~\cite{yang2024test, yasarla2025roca, lu2024garmentlab,
ren2023autonomous}. Retraining for every unseen condition is
prohibitively costly, motivating \emph{test-time adaptation}
(TTA)~\cite{wang2021tent, yuan2023robust, hoang2024persistent},
which adapts models during inference using only unlabeled test data
without revisiting the source training set.

TTA research has progressed through increasingly realistic
protocols. Early methods~\cite{wang2021tent, mummadi2021test}
assumed a single fixed target domain, while
CoTTA~\cite{wang2022continual} extended this to \emph{continual}
TTA (CTTA) with evolving domains under i.i.d.\ sampling.
\emph{Practical} TTA (PTTA)~\cite{yuan2023robust} further
introduces temporally correlated, non-i.i.d.\ sampling---the most
demanding and realistic setting. Under PTTA, each mini-batch covers
only a narrow slice of the target distribution, making
\emph{memory}---the mechanism that accumulates informative samples
over time---indispensable for stable
adaptation~\cite{yuan2023robust, hoang2024persistent, kang2024membn}.

Existing memory-based methods employ various selection criteria
(\eg, prediction entropy, class balancing, heuristic scoring) to
populate a single unstructured pool, which we term single-cluster
memory (SCM). While SCM alleviates the narrowness of pure in-batch
adaptation, it implicitly assumes that the target distribution is
unimodal. This raises a fundamental question: \emph{can such an
unstructured design faithfully represent the complex distributions
encountered during practical test-time adaptation?}

\begin{figure}[t]
    \centering
    \includegraphics[width=\linewidth]{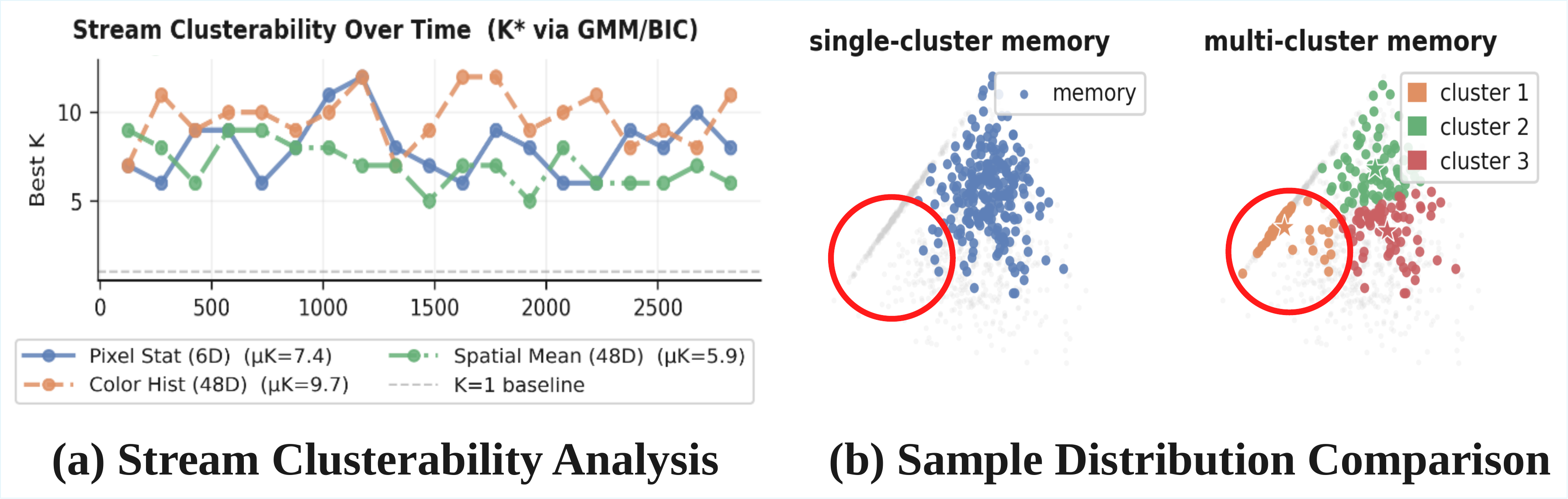}
    \caption{Motivation for multi-cluster memory. \textbf{(a)}~Stream clusterability analysis on CIFAR-100-C (PTTA): we fit GMMs with varying $K$ to sliding windows of the test stream and select the optimal $K^*$ via BIC across three descriptor types. The consistently high $K^*$ values ($\mu_{K^*}$ = 5.9--9.7) confirm that the target distribution is inherently \textit{multi-modal}, far exceeding the $K\!=\!1$ assumption of single-cluster memory. \textbf{(b)}~Under the same total capacity, SCM samples concentrate around similar regions of the descriptor space, whereas MCM distributes samples across distinct modes, extending coverage to under-represented regions (highlighted in red circle).}
    \label{fig:teaser}
\end{figure}

To answer this, we conduct a \emph{stream clusterability analysis}.
We collect sample descriptors within sliding windows of the test
stream and fit Gaussian Mixture Models (GMMs) with varying numbers
of components~$K$, evaluating three descriptor types: pixel-level
channel statistics, spatial mean, and color histograms. As shown in
Fig.~\ref{fig:teaser}(a), the Bayesian Information Criterion (BIC)
consistently selects $K \gg 1$ ($\mu_{K^{*}}$ ranging from $5.9$ to
$9.7$), providing strong evidence that streams are inherently
multi-modal even within a single corruption type. These results
establish that SCM is structurally insufficient for PTTA.

Motivated by this finding, we propose \textbf{Multi-Cluster Memory
(MCM)}, a structured memory framework that explicitly organizes
samples into multiple clusters based on lightweight pixel-level
statistical descriptors. MCM manages the memory lifecycle through
three mechanisms: \emph{descriptor-based cluster assignment} upon
sample arrival to capture distinct distributional modes,
\emph{Adjacent Cluster Consolidation} (ACC) to bound memory usage
by merging the most similar temporally adjacent clusters, and
\emph{Uniform Cluster Retrieval} (UCR) to ensure balanced
supervision across all modes during adaptation. As illustrated in
Fig.~\ref{fig:teaser}(b), MCM achieves substantially broader
coverage than SCM under the same capacity. MCM is a
\emph{plug-and-play} module compatible with any existing
memory-based TTA method.

We evaluate MCM on CIFAR-10-C,
CIFAR-100-C~\cite{krizhevsky2009learning},
ImageNet-C~\cite{hendrycks2019benchmarking}, and
DomainNet~\cite{peng2019moment} under the PTTA protocol. When
integrated with RoTTA~\cite{yuan2023robust},
PeTTA~\cite{hoang2024persistent}, and
ResiTTA~\cite{zhou2025resilient}, MCM yields consistent
improvements across all 12 baseline--dataset configurations, with
an average error reduction of 2.96\% and gains up to 5.00\% on
ImageNet-C and 12.13\% on DomainNet. Beyond accuracy, we introduce
a GMM-based diagnostic framework to directly assess \emph{memory
quality}---measuring imbalance ratio, distributional entropy, and
mode coverage against the evolving stream. This analysis reveals
that MCM maintains near-optimal balance and full mode coverage
throughout adaptation, whereas SCM exhibits persistent skew and
periodic mode loss, establishing a principled link between memory
representativeness and downstream adaptation performance. 

Our contributions are:
\begin{itemize}
    \item We present the first empirical evidence that PTTA data
    streams are inherently multi-modal, establishing a principled
    motivation for moving beyond single-cluster memory design.
    \item We propose Multi-Cluster Memory, a framework that organizes samples via pixel-level descriptor
    clustering, Adjacent Cluster Consolidation, and Uniform
    Cluster Retrieval, applicable to any memory-based TTA method.
    \item We introduce a GMM-based diagnostic framework that
    measures memory quality through imbalance ratio,
    entropy, and coverage, and demonstrate consistent
    improvements across benchmarks and baselines,
    confirming that gains stem from structured organization
    rather than increased capacity.
\end{itemize}

\section{Related Work}

\noindent\textbf{Evolution of Test-Time Adaptation Settings.}
Test-time adaptation (TTA) emerged as a paradigm for adapting
pre-trained models to target domains during inference without
access to source data. Early methods~\cite{mummadi2021test,
wang2021tent} operated under a \emph{fully TTA} setting, where
the entire test set originates from a single fixed target domain.
In this setup, all corruptions are treated uniformly, and
adaptation proceeds directly from the source-trained model without
accounting for temporal variation or domain evolution. Subsequently,
CoTTA~\cite{wang2022continual} extended this formulation to
\emph{continual TTA} (CTTA), where models must adapt to a sequence
of evolving domains. To mitigate catastrophic forgetting, CoTTA
introduces a stochastic restoration mechanism that intermittently
resets the model to its source-pretrained state. Follow-up
works~\cite{zhu2024reshaping, liu2024vida, han2025ranked} have
further advanced CTTA through improved regularization and
domain-aware strategies.

To better approximate real-world conditions, where samples from
consecutive time steps exhibit inherent correlations,
LAME~\cite{boudiaf2022parameter} was among the first to explicitly
address non-i.i.d.\ sampling. RoTTA~\cite{yuan2023robust}
further unified this with continual domain shift, giving rise to
the \emph{Practical TTA} (PTTA) paradigm that more faithfully
mirrors deployment scenarios. Building on this,
PeTTA~\cite{hoang2024persistent} introduced \emph{recurring TTA},
revealing that repeated adaptation cycles can eventually drive
models toward collapse. In this work, we focus primarily on PTTA
as the most demanding setting, with recurring TTA evaluations
provided in the appendix.

\noindent\textbf{Memory-Based TTA Systems.}
Memory has long served as a mechanism for preserving informative
samples in adaptive systems. In the context of TTA, memory can be
broadly categorized into explicit sample
storage~\cite{rolnick2019experience, song2023ecotta}, implicit
parametric memory~\cite{wu2022memvit, omidi2025memory,
tseng2025memory}, and retrieval from external
knowledge~\cite{lewis2020retrieval, wang2024searching}. Among
these, explicit memory has become indispensable for ensuring
stability and mitigating catastrophic forgetting under practical
TTA conditions.

Most existing methods rely on what we term single-cluster memory
(SCM), treating all stored samples as a homogeneous pool.
RoTTA~\cite{yuan2023robust} employs heuristic scoring based on
sample age and prediction uncertainty to manage memory turnover.
PeTTA~\cite{hoang2024persistent} extends this with persistent
strategies that maintain memory across recurring domain shifts.
ECoTTA~\cite{song2023ecotta} proposes self-distilled
regularization to prevent model drift, MemBN~\cite{kang2024membn}
focuses on maintaining reliable batch normalization statistics,
and ResiTTA~\cite{zhou2025resilient} introduces residual
connections to enhance robustness against normalization
degradation. Despite their individual contributions, all these
methods manage memory as a flat pool without awareness of the
underlying distributional modes, offering only a coarse
approximation of multi-modal target domains---a limitation that
becomes particularly pronounced under PTTA's non-i.i.d.\ sampling.

\noindent\textbf{Multi-Prototype and Structured TTA.}
Prior TTA methods have explored multi-prototype or structured
representations, but their designs serve fundamentally different
goals. AdaContrast~\cite{chen2022contrastive} maintains a
per-sample memory queue for contrastive pseudo-label refinement
via nearest-neighbor soft voting in the learned feature space.
SoTTA~\cite{gong2023sotta} uses class-level centers to filter
noisy or out-of-distribution samples through high-confidence
uniform-class sampling. While these methods improve prediction
quality within the classifier's learned representations, they
operate primarily in the semantic feature space and do not
explicitly model the multi-modal structure of domain shifts
arising from different corruption types or environmental conditions.

In contrast, MCM shifts the focus from semantic refinement to
\textbf{distributional simulation}. By organizing memory using
low-level pixel descriptors---channel-wise mean and variance that
directly reflect domain characteristics such as illumination,
noise, and blur---MCM captures the diverse appearance modes
inherent in PTTA streams. Rather than focusing on class
boundaries, our design ensures balanced and broad coverage of
the target manifold, mitigating the distributional bias inherent
in conventional SCM designs. As we demonstrate empirically, this
distributional perspective yields consistent gains when integrated
with existing memory-based TTA methods.

\section{Methodology}\label{sec:methodology}

\begin{figure}[t]
    \centering
    \includegraphics[width=\linewidth]{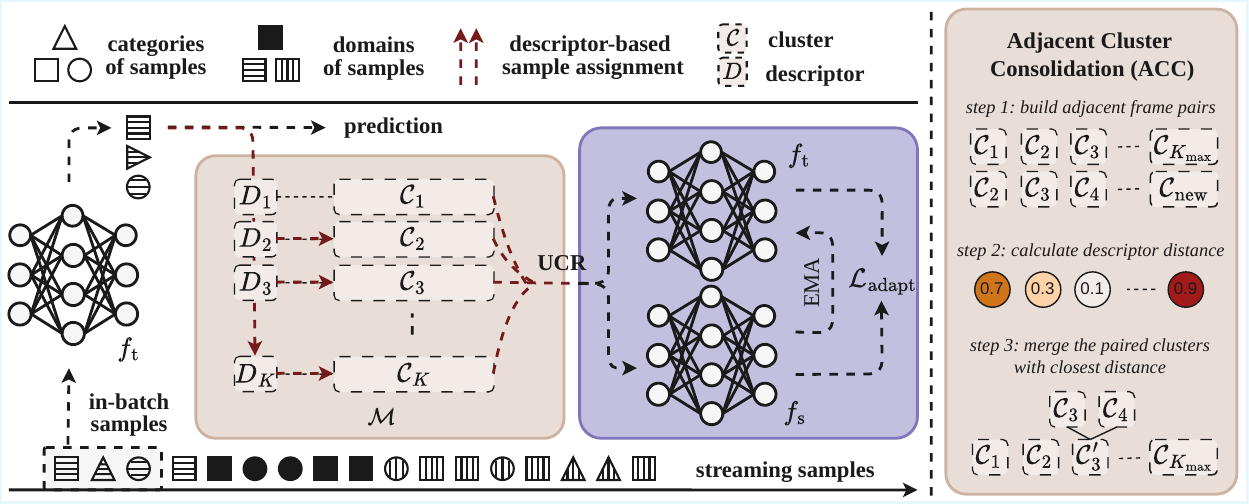}
    \caption{Overview of the TTA system with \methodname{} (MCM). Incoming samples are assigned to clusters via pixel-level descriptors (\textit{left}). Uniform Cluster Retrieval (UCR) draws balanced samples across all clusters for adaptation (\textit{center}). Adjacent Cluster Consolidation (ACC) merges the closest temporally adjacent pair when capacity is reached (\textit{right}). The three stages jointly preserve the multi-modal structure of the target stream under bounded memory.}
    \label{fig:method}
\end{figure}

\subsection{Revisiting Memory-based Test-Time Adaptation}

Current memory-based TTA approaches~\cite{yuan2023robust, hoang2024persistent} typically employ a single-cluster memory $\mathcal{M} = \{x_i\}_{i=1}^{N}$ that stores samples selected via criteria such as prediction entropy, class balancing, or heuristic scoring. At each time step $t$, the model receives a mini-batch $\mathcal{B}_t$ and first evaluates the outputs of each sample. Based on these outputs, the memory bank is updated with informative samples, and a subset is retrieved from $\mathcal{M}$ to adapt the model by minimizing
\begin{equation}
\label{eq:adapt}
  \mathcal{L}_{\mathrm{adapt}}
  = \mathbb{E}_{x \sim \mathcal{M}}
    \bigl[\mathcal{L}_{\mathrm{cons}}(f_{\mathrm{s}}(x),\, f_{\mathrm{t}}(x))\bigr].
\end{equation}
Following the Mean Teacher paradigm~\cite{tarvainen2017mean}, $f_{\mathrm{s}}$ and $f_{\mathrm{t}}$ denote the student and teacher networks, respectively. The teacher network produces pseudo-labels and is updated via an exponential moving average~(EMA).

While this design alleviates the distributional narrowness inherent to pure in-batch adaptation, it treats all stored samples as a \emph{homogeneous, unstructured} pool---an implicit assumption that the target distribution is unimodal. As demonstrated in Sec.~\ref{sec:introduction} (Fig.~\ref{fig:teaser}a), however, PTTA streams are inherently multi-modal: even within a single corruption type, the stream exhibits multiple distinct modes arising from visual diversity across semantic categories under non-i.i.d.\ sampling. A single-cluster memory has no mechanism to preserve this modal structure; its selection criteria operate on individual samples independently, without awareness of the underlying distributional modes. Consequently, the memory provides limited coverage of the descriptor space compared to a structured alternative, as illustrated in Fig.~\ref{fig:teaser}(b). We provide further quantitative evidence in Sec.~\ref{sec:experiments}, where SCM consistently exhibits higher imbalance ratios and lower distributional coverage than our proposed MCM.

This representativeness gap directly limits adaptation quality: the model is trained on a skewed sample pool that fails to reflect the full diversity of the incoming stream, reducing its ability to generalize across all modes of the target distribution. This observation motivates a structured memory design that explicitly preserves multi-modal diversity---the subject of the following sections.

\subsection{Test-time Adaptation System with Multi-Cluster Memory}
\label{subsec:mcm}

To address the representativeness gap identified above, we propose \textbf{Multi-Cluster Memory (MCM)}, which partitions the memory bank $\mathcal{M}$ into up to $K_{\max}$ clusters $\mathcal{C}_1, \mathcal{C}_2, \ldots, \mathcal{C}_K$ ($K \in \{1, 2, \ldots, K_{\max}\}$), starting from an empty state ($K=0$). MCM manages the memory through three stages that mirror the lifecycle of each sample: \emph{descriptor-based assignment} upon arrival, \emph{Adjacent Cluster Consolidation} when capacity is exceeded (Sec.~\ref{subsec:acc}), and \emph{Uniform Cluster Retrieval} during adaptation (Sec.~\ref{subsec:ucr}). We detail the first stage below.

Each cluster $\mathcal{C}_k$ maintains up to $N$ samples, ensuring a maximum total capacity of $K_{\max} \times N$. To efficiently manage cluster assignment and consolidation, we characterize each sample $x$ by its channel-wise statistics descriptor:
\begin{equation}
  d_x = [\mu_x^{(1)}, \sigma_x^{(1)}, \ldots, \mu_x^{(c)}, \sigma_x^{(c)}],
\end{equation}
where $\mu_x^{(c)}$ and $\sigma_x^{(c)}$ denote the mean and variance of the $c$-th channel computed across spatial dimensions $H \times W$ of the raw image. Following previous work in test-time normalization~\cite{tomar2024unmixing}, these channel-wise statistics effectively capture domain shift characteristics while maintaining computational efficiency. Each cluster $\mathcal{C}_k$ is summarized by its centroid descriptor $D_k$, computed as the average of all member descriptors:
\begin{equation}
\label{eq:clus_descript}
  D_k = \frac{1}{|\mathcal{C}_k|} \sum_{x \in \mathcal{C}_k} d_x.
\end{equation}

\textbf{Sample Assignment.} Upon arrival of a sample $x_t$ at time $t$, we compute the Euclidean distance between its descriptor $d_{x_t}$ and all existing cluster centroids:
\begin{equation}
  k^* = \operatorname*{arg\,min}_{k \in \{1,\ldots,K\}} \|d_{x_t} - D_k\|_2.
\end{equation}
If the minimum distance exceeds a threshold $\tau$, a new cluster is spawned as $\mathcal{C}_{K+1} = \{x_t\}$; otherwise, $x_t$ is assigned to the nearest cluster $\mathcal{C}_{k^*}$. The threshold $\tau$ governs the granularity of mode separation: smaller values yield finer partitioning that distinguishes subtle distributional variations, whereas larger values absorb greater diversity within each cluster. Because the descriptor operates on channel-wise pixel statistics whose scale is bounded and consistent across datasets, $\tau$ can be set once per descriptor type without per-dataset tuning. We empirically verify this robustness in Supplementary Materials.

\textbf{Sample Replacement.} When the target cluster $\mathcal{C}_{k^*}$ reaches capacity ($|\mathcal{C}_{k^*}| = N$), we employ a heuristic scoring function to identify the least valuable sample for replacement. Building upon the scoring function from~\cite{yuan2023robust}, we incorporate descriptor distance as an additional term:
\begin{equation}
  H(x) = \lambda_t \cdot \frac{1}{1+\exp(-A_x/N)} + \lambda_u \cdot \frac{U_x}{\log N_C} + \lambda_d \cdot \|d_x - D_{k^*}\|_2,
\end{equation}
where $A_x$ denotes the age of sample $x$ (i.e., the number of steps since insertion), $U_x$ represents its prediction entropy, $N_C$ is the number of classes, and $\lambda_t$, $\lambda_u$, $\lambda_d$ are reweighting coefficients that balance timeliness, uncertainty, and descriptor proximity, respectively. The sample with the highest score is replaced by $x_t$. By extending RoTTA's timeliness and uncertainty criteria with a descriptor distance term, this strategy ensures that clusters preserve not only temporal relevance and prediction confidence but also intra-cluster compactness.

\subsection{Adjacent Cluster Consolidation (ACC)}
\label{subsec:acc}
When the number of clusters reaches $K_{\max}$, a consolidation step is triggered to free capacity while preserving the multi-modal coverage of the memory. The key design choice is which pair of clusters to merge. Inspired by the adjacent-frame merging strategy in long-video memory management~\cite{song2024moviechat}, we restrict candidates to adjacent cluster pairs in the creation sequence. Under PTTA's temporally correlated streams, consecutively created clusters are likely to originate from similar or transitioning domains, making them natural merge candidates whose consolidation least disrupts the overall distributional coverage. This restriction also reduces the search from $O(K^2)$ pairwise comparisons to $O(K)$, yielding a favorable efficiency--quality trade-off (see comparison with alternative strategies in Sec.~\ref{sec:experiments}). For each adjacent pair $(\mathcal{C}_i, \mathcal{C}_{i+1})$, we compute their centroid distance $\Delta_{i,i+1} = \|D_i - D_{i+1}\|_2$ and merge the pair with the minimum distance. The consolidation process unifies all samples from both clusters into a single pool and retains the $N$ samples with the lowest prediction uncertainty. Formally, the merged cluster is defined as
\begin{equation}
\mathcal{C}_\text{merged} = \begin{cases}
\mathcal{C}_i \cup \mathcal{C}_{i+1} & \text{if } |\mathcal{C}_i \cup \mathcal{C}_{i+1}| \leq N \\
\text{top-}K(\{x \in \mathcal{C}_i \cup \mathcal{C}_{i+1} : U_x \text{ ascending}\},N) & \text{otherwise}
\end{cases}.
\end{equation}
Since the merged pair is the most similar adjacent clusters, their samples already share similar distributional characteristics; retaining the lowest-uncertainty subset thus prioritizes pseudo-label reliability without sacrificing cross-cluster diversity. The merged cluster descriptor is then recomputed via Eq.~(\ref{eq:clus_descript}).

\subsection{Uniform Cluster Retrieval (UCR)}
\label{subsec:ucr}
During adaptation, we retrieve an equal number of samples from every cluster in the memory bank. Concretely, given $K$ active clusters each storing up to $N$ samples, we draw $\lfloor N_{\text{adapt}} / K \rfloor$ samples from each cluster to form the retrieval set $\mathcal{M}_{\text{retrieve}}$, where $N_{\text{adapt}}$ denotes the total number of samples used per adaptation step. This uniform strategy is a direct consequence of our multi-modal motivation: since each cluster captures a distinct mode of the target distribution, equal representation during adaptation prevents the gradient signal from being dominated by any single mode. Following established TTA practices~\cite{yuan2023robust, hoang2024persistent}, we employ the Mean Teacher framework~\cite{tarvainen2017mean} with consistency regularization as defined in Eq.~\eqref{eq:adapt}. The novelty lies not in the adaptation mechanism, but in the composition of $\mathcal{M}_{\text{retrieve}}$: by construction, it mirrors the multi-modal structure of the target stream, ensuring that the model receives balanced supervision across all distributional modes at every adaptation step. In contrast, single-cluster memory draws all adaptation samples from a single undifferentiated pool, where dominant modes inevitably receive disproportionate representation.

\section{Experiments}\label{sec:experiments}

\subsection{Setup and Protocols}
\paragraph{Datasets and Metrics.}
We evaluate our method on four benchmark datasets under the
Practical Test-Time Adaptation (PTTA) setting~\cite{yuan2023robust}. For CIFAR10-C, CIFAR100-C, and ImageNet-C~\cite{hendrycks2019robustness}, we adopt corruption severity level~5, covering all 15 corruption types presented as a single temporally correlated stream.
We further evaluate on DomainNet~\cite{peng2019moment}, using 126 categories with \emph{real} as the source domain and \emph{clipart}, \emph{painting}, and \emph{sketch} as three separate target domains. Following prior work~\cite{yuan2023robust}, performance is measured by the mean classification error rate~(\%) averaged over all corruption types (or target domains for DomainNet).

\paragraph{Implementation Details.} 
All experiments are conducted on a single NVIDIA RTX 4090 GPU under the PTTA protocol of~\cite{yuan2023robust}. For MCM-specific hyperparameters, we set the per-cluster capacity $N=64$ and the descriptor distance threshold $\tau=0.3$. The maximum number of clusters is determined by $K_\text{max}=\min\!\bigl(5,\,\max(1,\,\lfloor N_c/20\rfloor)\bigr)$, where $N_c$ denotes the number of semantic classes in the dataset, yielding $K_\text{max}=1$ for CIFAR10-C and $K_\text{max}=5$ for CIFAR100-C, ImageNet-C, and DomainNet. This design reflects the observation that the degree of multi-modality in PTTA streams scales with the label space: a richer set of semantic categories produces greater visual diversity.

\paragraph{Baselines.} 
We compare against a comprehensive set of methods: Source (no adaptation), BN~\cite{nado2020evaluating}, PL~\cite{lee2013pseudo}, TENT~\cite{wang2021tent}, LAME~\cite{boudiaf2022parameter}, CoTTA~\cite{wang2022continual}, NOTE~\cite{gong2022note}, RDumb~\cite{press2023rdumb}, ROID~\cite{marsden2024universal}, TRIBE~\cite{Su_Xu_Jia_2024}, and NEO~\cite{murphy2025neo}. For all baselines with publicly available code, we use the official implementations with their default hyperparameters and the RobustBench~\cite{CroceEtAl2021_RobustBench} preprocessing pipeline to ensure a fair comparison. Results marked with $\dagger$ in the tables denote our own re-implementation when official results on a particular dataset were not reported.

\begin{table}[t]
\centering
\small
\caption{Practical Test-time Adaptation (PTTA) error rates (\%) on CIFAR10-C, CIFAR100-C, ImageNet-C, and DomainNet (severity 5). Lower is better. MCM is integrated into three memory-based baselines (RoTTA, PeTTA, ResiTTA) and yields consistent improvements across all configurations, with an average error reduction of 2.96\%. Numbers in parentheses indicate improvement over the respective baseline. $\dagger$~denotes our re-implementation. ``--'' indicates the method was not evaluated on that dataset in the original publication.}
\label{tab:ptta_summary}
\begin{tabular}{l|l|cccc}
\toprule
Method & Venue & CIFAR10-C & CIFAR100-C & ImageNet-C & DomainNet \\
\midrule
Source  & --          & 43.50 & 46.40 & 82.00 & --    \\
BN      & CoRR'20     & 75.20 & 52.90 & --    & --    \\
PL      & ICML'13     & 82.90 & 88.90 & --    & --    \\
TENT    & ICLR'21     & 86.00 & 92.80 & --    & --    \\
LAME    & CVPR'22     & 39.50 & 40.50 & 80.90 & --    \\
CoTTA   & CVPR'22     & 83.20 & 52.20 & 98.60 & --    \\
NOTE    & NeurIPS'22  & 31.10 & 73.80 & --    & --    \\
RDumb   & NeurIPS'23  & 31.10 & 36.70 & 72.20 & 44.30 \\
ROID    & WACV'24     & 72.70 & 76.40 & 62.70 & --    \\
TRIBE   & AAAI'24     & \textbf{15.30} & 33.80 & 63.60 & --    \\
NEO     & ICLR'26     & 46.36$^\dagger$ & 43.25$^\dagger$ & 78.25$^\dagger$ & --    \\
\midrule
RoTTA   & CVPR'23     & 25.20 & 35.00 & 68.30 & 44.30 \\
\quad + MCM & --      & 22.59 (\textcolor{green!60!black}{-2.61}) & 33.75 (\textcolor{green!60!black}{-1.25}) & 67.46 (\textcolor{green!60!black}{-0.84}) & 42.53 (\textcolor{green!60!black}{-1.77}) \\
\midrule
PeTTA   & NeurIPS'24  & 24.30 & 35.80 & 65.30 & 43.80 \\
\quad + MCM & --      & 21.55 (\textcolor{green!60!black}{-2.75}) & 33.04 (\textcolor{green!60!black}{-2.76}) & \textbf{60.30} (\textcolor{green!60!black}{-5.00}) & 42.80 (\textcolor{green!60!black}{-1.00}) \\
\midrule
ResiTTA & ICASSP'25   & 22.80 & 32.50 & 69.40 & 54.76$^\dagger$ \\
\quad + MCM & --      & \underline{20.69} (\textcolor{green!60!black}{-2.11}) & \textbf{31.90} (\textcolor{green!60!black}{-0.60}) & 66.65 (\textcolor{green!60!black}{-2.75}) & \textbf{42.63} (\textcolor{green!60!black}{-12.13}) \\
\bottomrule
\end{tabular}
\end{table}

\subsection{Main Results}
Table~\ref{tab:ptta_summary} reports error rates under the PTTA setting across benchmarks. We integrate MCM into three representative memory-based TTA methods, \eg, RoTTA, PeTTA, and ResiTTA, to evaluate both effectiveness and generalizability. Notably, NEO---the current state-of-the-art under the CTTA protocol---suffers substantial degradation when evaluated under PTTA, underscoring the additional difficulty posed by temporally correlated, non-i.i.d.\ sampling.

\paragraph{Consistent improvements across baselines.}
MCM reduces the error rate for every baseline on every dataset, achieving an average improvement of 2.96\% across all 12 baseline--dataset configurations. The largest single gains are observed with PeTTA on ImageNet-C (\textbf{60.30\%}, $-5.00$\%) and with ResiTTA on DomainNet (\textbf{42.63\%}, $-12.13$\%). MCM also achieves the best overall result on CIFAR100-C (\textbf{31.90\%}) when combined with ResiTTA. The notably large gain on DomainNet with ResiTTA is partly attributable to the baseline's reliance on batch-normalization statistics, which degrade under the severe domain gap of the real$\,\to\,$sketch/painting/clipart transfer; MCM's multi-cluster structure provides more representative statistics for adaptation, substantially alleviating this issue.

\paragraph{Scaling with distributional complexity.}
The benefits of MCM scale with the complexity of the target distribution. On ImageNet-C and DomainNet, which exhibit larger label spaces and consequently higher multi-modality under non-i.i.d.\ sampling, the average improvements reach 2.86\% and 4.97\%, respectively. In contrast, CIFAR10-C contains only 10 classes, limiting the degree of multi-modality in the stream and thus the headroom for multi-cluster organization; nonetheless, MCM still provides consistent gains of 2.11--2.75\% over all three baselines. We note that TRIBE achieves 15.30\% on CIFAR10-C via a tri-net self-training architecture with balanced batch normalization, which adapts directly on the current mini-batch without maintaining a memory bank. As MCM is a memory-side module, it is not directly applicable to memory-free methods such as TRIBE; the two approaches address complementary aspects of TTA.

\subsection{Ablation Study}

\begin{table}[t]
\centering
\small
\setlength{\tabcolsep}{4pt}
\begin{minipage}[t]{0.46\linewidth}
\centering
\captionof{table}{Component ablation. Each module
contributes independently; combining all three
yields the lowest error.}
\label{tab:component}
\begin{tabular}{ccc|c}
\toprule
MCM & ACC & UCR & Error (\%) \\
\midrule
            &             &             & 35.80 \\
\checkmark  &             &             & 35.33 \\
\checkmark  & \checkmark  &             & 34.09 \\
\checkmark  &             & \checkmark  & 35.33 \\
\checkmark  & \checkmark  & \checkmark  & \textbf{33.04} \\
\bottomrule
\end{tabular}
\end{minipage}
\hfill
\begin{minipage}[t]{0.50\linewidth}
\centering
\captionof{table}{Descriptor and threshold
sensitivity. Pixel-level statistics consistently
outperform CNN features by a large margin.}
\label{tab:descriptor}
\begin{tabular}{l|cc}
\toprule
Descriptor ($\tau$) & Error (\%) & Time (s) \\
\midrule
CNN Feature (5.0) & 43.30 & 827.67 \\
\midrule
Pixel Stat (0.1) & 35.31 & \textbf{987.71} \\
Pixel Stat (0.3) & \textbf{33.04} & 1049.37 \\
Pixel Stat (0.5) & 35.59 & 1765.16 \\
Pixel Stat (0.7) & 36.85 & 1349.07 \\
\bottomrule
\end{tabular}
\end{minipage}
\end{table}

In this section, we conduct extensive analyses to justify the design choices of MCM. Unless otherwise specified, all ablations are conducted on CIFAR-100-C (PTTA) with PeTTA~\cite{hoang2024persistent} as the base method.

\paragraph{Component ablation.}
Table~\ref{tab:component} isolates the contribution of each MCM component. Starting from PeTTA with its default single-cluster memory (35.80\%), replacing it with a multi-cluster structure alone yields 35.33\%. Adding ACC further reduces the error to 34.09\%, confirming that temporally aware consolidation better preserves distributional diversity. Applying UCR alone on top of the multi-cluster structure also yields 35.33\%, showing that balanced retrieval independently benefits adaptation. The full model achieves the lowest error of 33.04\%, demonstrating clear synergy: ACC maintains diverse modes in storage, while UCR ensures each mode contributes equally during learning.

\paragraph{Descriptor space and threshold sensitivity.}
Table~\ref{tab:descriptor} compares pixel-level and CNN-feature descriptors. Even at its best threshold ($\tau\!=\!5.0$, selected from seven candidates), the feature-based descriptor only reaches 43.30\%, whereas the pixel-based descriptor achieves 33.04\% at $\tau\!=\!0.3$---a gap of over 10 percentage points. We attribute this to the clustering objective in MCM: CNN features are trained to suppress low-level variations and amplify semantic differences, so feature-based clusters group samples by class rather than by corruption type. Pixel-level channel statistics, in contrast, directly capture domain-shift characteristics such as illumination and noise level, aligning with MCM's goal of partitioning the stream by distributional mode.

Among pixel-descriptor thresholds, $\tau\!=\!0.3$ yields the best error rate while maintaining moderate runtime. Smaller values ($\tau\!=\!0.1$) create too many fine-grained clusters that are frequently consolidated, reducing effective diversity; larger values ($\tau\!\geq\!0.5$) absorb heterogeneous samples into the same cluster, diluting mode separation. The narrow spread across all four thresholds (33.04--36.85\%) confirms that the pixel descriptor is robust to this hyperparameter.

\begin{table}[t]
\centering
\small
\setlength{\tabcolsep}{4pt}
\begin{minipage}[t]{0.46\linewidth}
\centering
\captionof{table}{Comparison of cluster consolidation
strategies. ACC leverages temporal adjacency to merge
the most similar neighboring clusters, achieving the
best accuracy and efficiency. All strategies use the
same MCM framework with PeTTA on CIFAR-100-C (PTTA).}
\label{tab:consolidation}
\begin{tabular}{l|cc}
\toprule
Strategy & Error (\%) & Time (s) \\
\midrule
GCC  & 34.26 & 1094 \\
SCM  & 36.71 & 1665 \\
LRU  & 33.78 & 1271 \\
\midrule
ACC  & \textbf{33.04} & \textbf{1049} \\
\bottomrule
\end{tabular}
\end{minipage}
\hfill
\begin{minipage}[t]{0.50\linewidth}
\centering
\captionof{table}{Recurring TTA error rates (\%) on
CIFAR-100-C (20 rounds, severity 5). Lower is better.
MCM sustains long-term gains without catastrophic
forgetting.}
\label{tab:recurring}
\begin{tabular}{l|cc}
\toprule
Method & Round 1 & Avg (20) \\
\midrule
Source & 46.5 & 46.5 \\
CoTTA & 53.4 & 83.1 \\
RoTTA & 35.5 & 61.4 \\
RDumb & 36.7 & 36.6 \\
TRIBE & 33.8 & 39.6 \\
PeTTA & 35.8 & 35.1 \\
\midrule
PeTTA + MCM & \textbf{33.8} & \textbf{32.6} \\
\bottomrule
\end{tabular}
\end{minipage}
\end{table}

\paragraph{Consolidation strategies.}
When the number of clusters reaches $K_\text{max}$, a consolidation step is triggered. We compare four strategies (Table~\ref{tab:consolidation}): \emph{Adjacent Cluster Consolidation} (ACC) merges the closest pair among temporally adjacent clusters; \emph{Global Cluster Consolidation} (GCC) merges the globally closest pair in descriptor space; \emph{Smallest Cluster Merging} (SCM) merges the cluster with the fewest samples; and \emph{Least Recently Used} (LRU) merges the least active cluster. ACC achieves both the lowest error (33.04\%) and the fastest runtime (1049\,s). GCC reaches 34.26\%: although geometrically intuitive, it may merge clusters from distant time steps that cover complementary regions, inadvertently reducing coverage. SCM performs worst (36.71\%) because it systematically discards under-represented modes---precisely the minority modes MCM aims to preserve. LRU (33.78\%) is competitive but slower, as inactivity does not reliably indicate redundancy. These results confirm that temporal adjacency is an effective inductive bias under PTTA's correlated streams.

\subsection{Further Analysis}

\paragraph{Long-term stability under recurring TTA.}
We further evaluate MCM under the recurring TTA protocol
introduced by PeTTA~\cite{hoang2024persistent}, where the
model repeatedly traverses the same corruption sequence over
20 rounds---a stringent test of whether adaptation remains
stable or gradually collapses. As shown in
Table~\ref{tab:recurring}, TRIBE---which achieves the best
single-pass result on CIFAR10-C (15.30\%)---escalates from
33.8\% to 39.6\% average over 20 rounds, revealing that
strong initial performance does not guarantee long-term
stability. PeTTA maintains a stable 35.1\% average across
all rounds, yet shows no further improvement over time.
PeTTA+MCM not only preserves this stability but continues
to improve, reaching a 20-round average of 32.6\%
(Round~1: 33.8\% $\to$ Round~20: 32.5\%). We attribute this
to MCM's structured memory: built upon PeTTA's 
robust adaptation mechanism, the multi-cluster organization
accumulates increasingly representative samples across all
distributional modes as corruption types recur, translating
better memory quality into sustained performance gains.

\begin{figure}[t]
    \centering
    \includegraphics[width=1\linewidth]{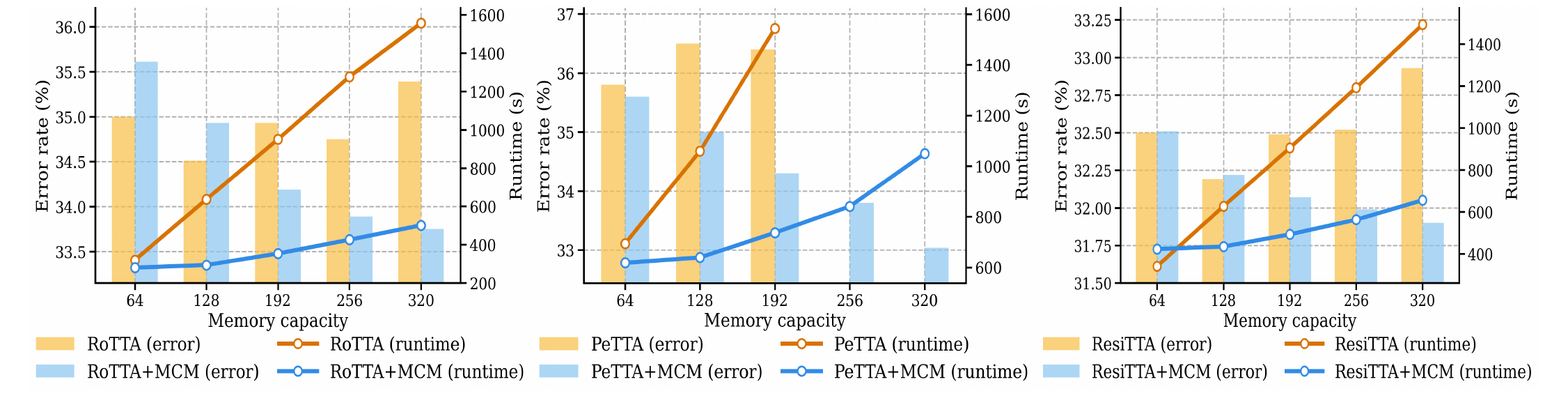}
    \caption{Memory scaling comparison on CIFAR-100-C (PTTA). Bars denote error rate; lines denote runtime. For MCM, per-cluster capacity is fixed at 64 and total capacity is varied by the number of clusters. Across all three baselines, simply enlarging the single-cluster pool increases runtime with negligible accuracy gain, whereas MCM consistently achieves lower error at lower cost under equal total capacity. PeTTA with SCM at 256 and 320 samples encountered out-of-memory errors (middle panel).}
    \label{fig:mem_sacling_runtime}
\end{figure}

\paragraph{Structured vs.\ unstructured memory scaling.}
To verify that MCM's gains stem from memory organization rather than increased capacity, we scale both single-cluster memory (SCM) and MCM to equal total sizes on CIFAR-100-C (Figure~\ref{fig:mem_sacling_runtime}). Enlarging SCM from 64 to 320 samples yields negligible accuracy improvement across all three baselines while incurring up to $5\times$ runtime growth, indicating that naively storing more samples fails to capture distributional diversity. In contrast, MCM structures the same 320 samples into 5 clusters of 64, consistently achieving lower error at lower runtime. For example, PeTTA+MCM at 320 samples reaches 33.04\% in 1049\,s, whereas PeTTA with SCM at 192 samples already exceeds both in error and runtime---and encounters out-of-memory errors beyond that point. These results confirm that the improvements of MCM arise from principled multi-cluster organization, not from raw storage capacity.

\begin{figure}[t]
    \centering
    \includegraphics[width=\linewidth]{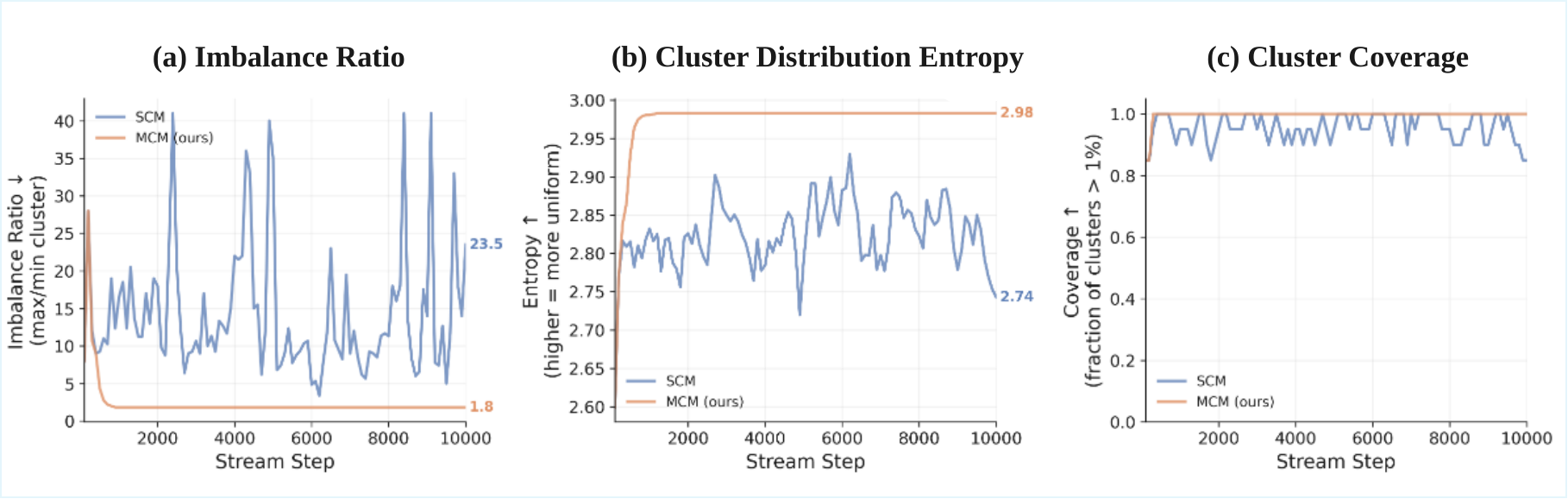}
    \caption{Diagnostic comparison of memory quality between SCM and MCM over the CIFAR-100-C stream (PTTA, PeTTA). We fit a GMM to the evolving stream and measure three properties of the stored memory: \textbf{(a)}~imbalance ratio (lower is better), \textbf{(b)}~distributional entropy (higher is more uniform), and \textbf{(c)}~cluster coverage (fraction of GMM components with $>$1\% representation). MCM maintains near-constant balance, entropy, and coverage throughout adaptation, whereas SCM exhibits high variance and progressive degradation.}
    \label{fig:memory_evaluation}
\end{figure}

\paragraph{Memory quality diagnostics.}
To understand \emph{why} MCM improves adaptation, we analyze the memory content itself. We fit a GMM to the evolving CIFAR-100-C stream as a reference distribution and track three diagnostic metrics over time
(Figure~\ref{fig:memory_evaluation}). \textbf{(a)}~The imbalance ratio measures the max-to-min cluster occupancy; SCM fluctuates between 10--40$\times$ and settles at 23.5, while MCM remains near 1.8 throughout, indicating  consistently balanced mode representation. \textbf{(b)}~Distributional entropy quantifies how uniformly samples are spread across GMM components; MCM sustains near-maximum entropy (2.98) whereas SCM drops to 2.74 with high variance, reflecting skewed coverage that shifts unpredictably over time. \textbf{(c)}~Mode coverage tracks the fraction of GMM components with meaningful representation ($>$1\%) in memory; MCM maintains full coverage ($\approx$1.0) while SCM periodically loses entire modes. Together, these results confirm that MCM's performance gains are rooted in superior distributional representativeness: by explicitly preserving multi-modal structure, MCM provides the adaptation process with a more faithful approximation of the target stream at every time step.

\section{Discussion}\label{sec:discussion}

Our study highlights a key but often overlooked design axis in
memory-based test-time adaptation: \emph{how memory is organized}.
While prior work largely focuses on increasing memory capacity
or refining sample selection, our results show that the main
bottleneck lies not in \emph{how many} samples are stored,
but in \emph{how} they are structured. Empirically, enlarging a single-cluster memory pool yields minimal gains while incurring substantial computational cost.
In contrast, organizing the same capacity into multiple
descriptor-based clusters consistently improves both adaptation
accuracy and runtime efficiency. Our GMM-based diagnostics
further support this finding: the improvements of MCM directly
correspond to better distributional balance, higher entropy,
and improved mode coverage.

The effectiveness of pixel-level descriptors also provides a
broader insight for representation design. Many existing
approaches rely on high-level CNN features to characterize
samples in memory; however, such representations are optimized
to suppress domain-specific variations. In corruption-based
benchmarks, domain shifts instead manifest in low-level image
statistics such as color distributions and channel-wise
intensity patterns. Modeling memory structure in this space
therefore provides a more faithful representation of
distributional changes, suggesting that descriptor design
may play a broader role across TTA components.

\paragraph{Limitations.}
Despite its effectiveness, MCM has several limitations.
First, it introduces additional overhead from descriptor
computation and cluster management. Although empirical
measurements (Table~\ref{tab:consolidation}) show that this
cost remains modest relative to the adaptation process,
it may become more noticeable when scaling to higher-resolution
inputs or larger models. Second, the descriptor relies on
channel-wise pixel statistics, implicitly assuming that
domain shifts appear in low-order image statistics; this
bias may be less effective for shifts driven by geometric
transformations or high-level semantic changes. Third, our
experiments focus on corruption-based image classification;
extending the proposed memory organization strategy to other
modalities (\eg, language or multi-modal systems) and tasks
(\eg, detection, segmentation) remains an open direction.

\paragraph{Future Work.}
Two directions appear promising. First, replacing raw sample
storage with compact or learned representations could reduce
memory footprint and enable memory-based TTA for large
foundation models and multi-modal systems. Second, while MCM
currently orders clusters by creation time, richer structures
may better capture evolving distributions. For instance,
graph-based cluster management could relate clusters through
descriptor distances rather than temporal adjacency, enabling
more flexible consolidation.

\section{Conclusion}

We introduced Multi-Cluster Memory (MCM), a structured memory
framework for test-time adaptation that organizes samples via
descriptor-based clustering, Adjacent Cluster Consolidation,
and Uniform Cluster Retrieval. A stream clusterability
analysis shows that PTTA data streams exhibit intrinsic
multi-modal structure, motivating the transition from
single-cluster to multi-cluster memory. Extensive experiments on CIFAR-10-C, CIFAR-100-C, ImageNet-C,
and DomainNet show that MCM consistently improves existing
memory-based TTA methods, with gains increasing alongside
the distributional complexity of the test stream. Recurring
TTA evaluations further demonstrate that these improvements
remain stable—and continue to grow—over 20 adaptation rounds.
A GMM-based diagnostic framework links these gains to improved
memory quality in terms of distributional balance, entropy,
and mode coverage. Overall, our results establish that memory
\emph{organization}—rather than memory \emph{capacity}—is a
fundamental design axis for effective test-time adaptation.

\clearpage
\bibliographystyle{splncs04}
\bibliography{main}

\clearpage
\appendix
\renewcommand{\thetable}{S\arabic{table}}
\renewcommand{\thefigure}{S\arabic{figure}}
\renewcommand{\theequation}{S\arabic{equation}}
\setcounter{table}{0}
\setcounter{figure}{0}
\setcounter{equation}{0}
\renewcommand{\thesection}{\Alph{section}}
\renewcommand{\thesubsection}{\Alph{section}.\arabic{subsection}}

\section*{Appendix Overview}
\addcontentsline{toc}{section}{Appendix Overview}

\noindent This appendix provides additional analyses 
and results organized as follows:

\begin{itemize}[leftmargin=1.5em, itemsep=2pt]
    \item \textbf{Sec.~\ref{appdx:threshold-robustness}}: 
          Distance threshold~$\tau$ sensitivity
    \item \textbf{Sec.~\ref{appdx:distance-ablation}}: 
          Distance metric ablation
    \item \textbf{Sec.~\ref{appdx:kmax-sensitivity}}: 
          Maximum cluster count~$K_{\max}$ sensitivity
    \item \textbf{Sec.~\ref{appdx:recurring}}: 
          Full per-round recurring TTA results (20 rounds)
    \item \textbf{Sec.~\ref{appdx:visualization}}: 
          Memory sample distribution visualization
\end{itemize}


\section{Distance Threshold Sensitivity}
\label{appdx:threshold-robustness}

As discussed in the main paper (Sec.~3.2), the channel-wise 
pixel-statistics descriptor operates on a bounded and consistent 
scale, allowing the threshold~$\tau$ to be set once without 
per-dataset tuning. Here we complement the single-pass PTTA 
ablation (main paper, Table~3) with a detailed sensitivity 
analysis under the more demanding recurring TTA protocol on 
CIFAR-100-C~\cite{krizhevsky2009learning}.

Table~\ref{tab:ablation-threshold} reports per-round error rates 
for four thresholds across 20 recurring rounds using 
PeTTA~\cite{hoang2024persistent}+MCM. The overall spread is 
narrow (33.1--34.9\%), confirming that MCM is robust to the 
choice of~$\tau$, with $\tau{=}0.3$ consistently achieving the 
lowest average error (33.1\%). Two complementary failure modes 
are visible at the extremes. A small threshold ($\tau{=}0.1$, 
Avg~34.5\%) over-fragments the descriptor space, spawning 
clusters for minor distributional variations that are then 
frequently merged by ACC, effectively erasing the diversity they 
were meant to capture. A large threshold ($\tau{=}0.7$, 
Avg~34.9\%) produces the opposite effect: heterogeneous samples 
are absorbed into the same cluster, conflating distinct modes 
and diluting the representativeness of each cluster centroid. 
The intermediate value $\tau{=}0.5$ (Avg~34.3\%) partially 
mitigates both issues but still under-separates certain 
corruption transitions compared to $\tau{=}0.3$.

Beyond average performance, $\tau{=}0.3$ is also the only 
setting that exhibits a consistent downward trend in error 
over rounds (33.3\% at Round~1 $\rightarrow$ 32.8\% at 
Round~20), despite minor fluctuations in intermediate rounds. This progressive improvement indicates that the 
multi-cluster structure at this granularity accumulates 
increasingly representative samples as corruption types recur, 
translating better memory quality into sustained adaptation 
gains. At other thresholds, error rates plateau after the first 
few rounds, suggesting that the memory composition stabilizes 
prematurely before reaching full distributional coverage.

\begin{table}[t]
\centering
\caption{\textbf{Sensitivity of the distance threshold~$\tau$ on CIFAR-100-C (recurring TTA, severity~5).} Columns 1--20 report per-round classification error rates~(\%); \textit{Avg} is the mean over all 20 rounds. All results use PeTTA+MCM. \textbf{Bold} indicates the best value per column.}
\label{tab:ablation-threshold}
\resizebox{\linewidth}{!}{%
\begin{tabular}{l|cccccccccccccccccccc|c}
\toprule
$\tau$ & 1 & 2 & 3 & 4 & 5 & 6 & 7 & 8 & 9 & 10 & 11 & 12 & 13 & 14 & 15 & 16 & 17 & 18 & 19 & 20 & Avg \\
\midrule
0.1 & 34.8 & 34.4 & 34.5 & 34.6 & 34.7 & 34.6 & 34.8 & 34.7 & 34.7 & 34.5 & 34.6 & 34.4 & 34.3 & 34.4 & 34.2 & 34.3 & 34.4 & 34.3 & 34.1 & 34.1 & 34.5 \\
\textbf{0.3} & \textbf{33.3} & \textbf{33.1} & \textbf{33.3} & \textbf{33.3} & \textbf{33.4} & \textbf{33.2} & \textbf{33.2} & \textbf{33.2} & \textbf{33.0} & \textbf{33.0} & \textbf{33.0} & \textbf{33.1} & \textbf{33.0} & \textbf{33.0} & \textbf{33.0} & \textbf{32.9} & \textbf{32.9} & \textbf{32.9} & \textbf{32.9} & \textbf{32.8} & \textbf{33.1} \\
0.5 & 35.1 & 34.4 & 34.4 & 34.5 & 34.4 & 34.4 & 34.4 & 34.4 & 34.1 & 34.2 & 34.1 & 34.2 & 34.0 & 34.1 & 34.2 & 34.3 & 34.3 & 34.1 & 34.0 & 34.2 & 34.3 \\
0.7 & 35.5 & 34.6 & 34.9 & 35.0 & 35.0 & 34.9 & 34.9 & 34.8 & 34.9 & 34.9 & 34.8 & 34.9 & 34.8 & 34.7 & 34.9 & 35.0 & 34.8 & 34.8 & 34.8 & 34.7 & 34.9 \\
\bottomrule
\end{tabular}}
\end{table}

\section{Ablation on Distance Metrics}
\label{appdx:distance-ablation}

In the main paper, we adopt Euclidean distance for both cluster assignment (Eq.~4) and Adjacent Cluster Consolidation (Sec.~3.3). Here we ablate this choice by comparing distance metrics for descriptor-space computation. The experiment is conducted on CIFAR-100-C under the PTTA protocol using PeTTA~\cite{hoang2024persistent}+MCM. For each metric, we perform a grid search over its corresponding distance threshold~$\tau$ and report the best-performing configuration in Table~\ref{tab:distance_metrics}.

\begin{table}[t]
    \small
    \centering
    \caption{\textbf{Ablation on distance metrics for MCM descriptor-space computation.} Evaluated on CIFAR-100-C (PTTA, severity~5) with PeTTA+MCM. Each metric is paired with its best threshold from a grid search.}
    \label{tab:distance_metrics}
    \begin{tabular}{lcc}
    \toprule
    \textbf{Distance Metric} & \textbf{Best Threshold ($\tau$)} & \textbf{Avg.\ Error (\%)} \\
    \midrule
    Cosine & 0.05 & 34.27 \\
    Mahalanobis & 1.0 & 35.16 \\
    Manhattan ($L_1$) & 0.5 & 33.90 \\
    \midrule
    \textbf{Euclidean ($L_2$) (Ours)} & 0.3 & \textbf{33.04} \\
    \bottomrule
    \end{tabular}
\end{table}

\paragraph{Analysis.}
Euclidean distance ($L_2$) yields the lowest error (33.04\%) by providing an isotropic penalty that smoothly captures multidimensional domain shifts.
Manhattan distance ($L_1$) is competitive at 33.90\% but slightly less effective, likely because equal weighting of each coordinate does not account for the correlated nature of channel-wise mean--variance pairs.
Cosine distance performs worse (34.27\%) because it evaluates only the angle between descriptors, discarding magnitude information. While angular differences alone capture most inter-mode distinctions, magnitude provides additional discriminative power for separating modes with similar corruption profiles but differing intensities, accounting for the 1.2-point gap relative to Euclidean distance.
Mahalanobis distance yields the highest error (35.16\%) despite being widely used for feature-space domain matching. In our setting, the lightweight descriptors already explicitly contain channel-wise variance, so a secondary covariance normalization distorts the natural geometry of the descriptor space; moreover, estimating robust covariance matrices for dynamically forming micro-clusters in an online stream is numerically unstable.
These results confirm that Euclidean distance provides the best accuracy for MCM's pixel-level descriptors.

\section{Sensitivity of Maximum Cluster Count $K_{\max}$}
\label{appdx:kmax-sensitivity}

The main paper (Sec.~4.1) sets the maximum number of clusters 
via $K_{\max} = \min\!\big(5,\;\max(1,\lfloor N_c/20\rfloor)
\big)$, yielding $K_{\max}{=}5$ for CIFAR-100-C, ImageNet-C, 
and DomainNet, and $K_{\max}{=}1$ for CIFAR-10-C. Here we 
sweep $K_{\max}\in\{1,2,3,4,5,7,10\}$ on CIFAR-100-C (PTTA, 
severity~5) with PeTTA~\cite{hoang2024persistent}+MCM to 
justify this choice. The per-cluster capacity is fixed at 
$N{=}64$, so increasing $K_{\max}$ proportionally increases 
the total memory budget.

\begin{figure}[t]
    \centering
    \includegraphics[width=1.0\linewidth]{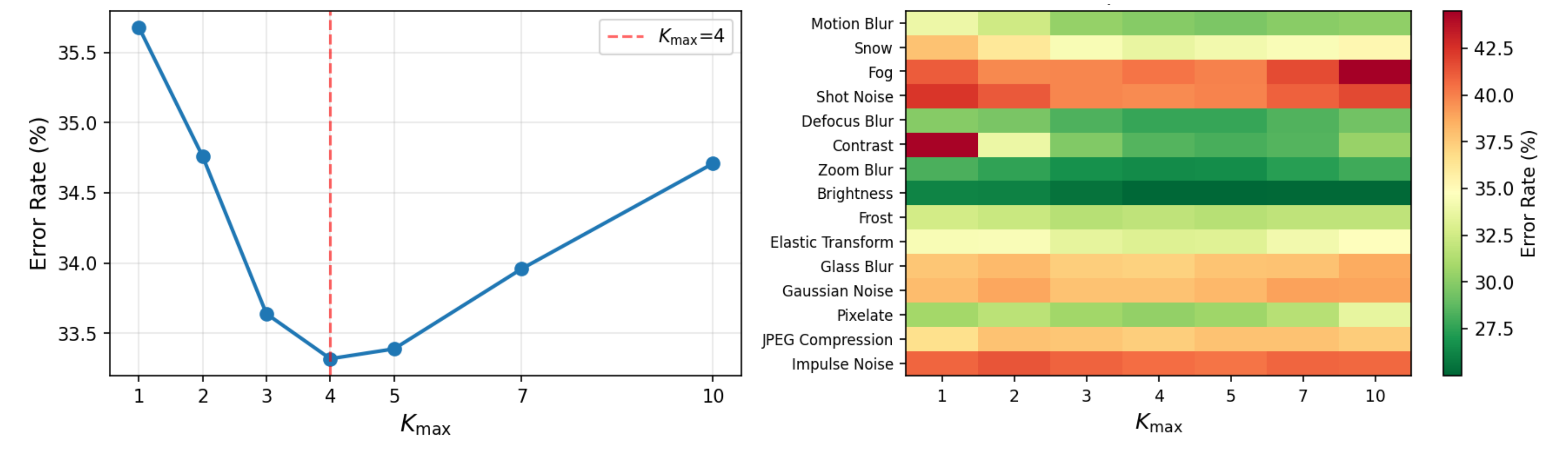}
    \caption{\textbf{Sensitivity of $K_{\max}$ on CIFAR-100-C 
    (PTTA, severity~5) with PeTTA+MCM.} \emph{Left}: average 
    error rate~(\%) vs.\ $K_{\max}$, showing a U-shaped trend 
    with minimum at $K_{\max}{=}4$ (dashed red line). 
    \emph{Right}: per-corruption error heatmap (green~=~lower 
    error). Most corruptions favor moderate $K_{\max}$ (3--5), 
    while contrast and impulse noise are most sensitive to this 
    parameter.}
    \label{fig:kmax_sensitivity}
\end{figure}

\paragraph{Overall trend.}
As shown in Fig.~\ref{fig:kmax_sensitivity} (left), the average 
error follows a U-shaped curve. $K_{\max}{=}1$ reduces to 
single-cluster memory and yields the highest error, confirming 
that multi-cluster organization is essential. Performance 
improves steadily from $K_{\max}{=}1$ to~4, where it reaches 
its minimum, with $K_{\max}{=}5$ performing nearly identically. 
Beyond this range, over-fragmentation degrades accuracy: too 
many clusters lead to small, statistically unstable populations 
and trigger frequent ACC merges that discard useful samples. 
The total spread remains moderate (${\sim}2.4$ percentage 
points), indicating that MCM is not overly sensitive to this 
hyperparameter.

\paragraph{Per-corruption analysis.}
The heatmap (Fig.~\ref{fig:kmax_sensitivity}, right) shows that 
most corruptions favor moderate cluster counts 
($K_{\max}\in\{3,4,5\}$) and degrade at both extremes. Two 
corruptions stand out as particularly sensitive: contrast and 
impulse noise exhibit substantially higher error at 
$K_{\max}{=}1$, as their pixel-level statistics diverge sharply 
from other corruption types and thus require dedicated clusters 
for adequate representation. Notably, impulse noise continues to 
improve even at $K_{\max}{=}10$, suggesting that its 
distributional modes are especially fine-grained; however, this 
per-corruption gain is outweighed by the degradation of other 
types at high $K_{\max}$. Corruptions that primarily alter 
spatial structure (e.g., motion blur, elastic transform) remain 
stable across all settings, as their channel statistics overlap 
with neighboring corruption types.

\paragraph{Justification.}
The near-identical performance at $K_{\max}{=}4$ and~5 confirms 
that the formula in the main paper operates within the optimal 
range. We retain $K_{\max}{=}5$ as the default because the 
difference is negligible and the extra cluster provides 
additional capacity for datasets with greater distributional 
complexity (ImageNet-C: 200 classes; DomainNet: 126 classes 
across heterogeneous styles). Coupling $K_{\max}$ to the label 
space via $\lfloor N_c/20\rfloor$ thus provides a tuning-free 
heuristic that adapts to dataset complexity without requiring 
per-dataset sweeps.

\section{Recurring TTA: Full Per-Round Results}
\label{appdx:recurring}

The main paper (Table~5) summarizes the recurring TTA experiment with Round~1 and 20-round average error rates. Here we provide the complete per-round breakdown on both CIFAR-100-C and CIFAR-10-C~\cite{krizhevsky2009learning} (Tables~\ref{tab:recurring-cifar100} and~\ref{tab:recurring-cifar10}), enabling a fine-grained view of how each method behaves as the model repeatedly traverses the same corruption sequence.

\paragraph{CIFAR-100-C (Table~\ref{tab:recurring-cifar100}).}
Several methods exhibit clear instability over extended adaptation. CoTTA~\cite{wang2022continual} and EATA~\cite{niu2022efficient} collapse rapidly, rising from ${\sim}50$--$88$\% at Round~1 to over 97\% by Round~20, confirming severe catastrophic forgetting. RoTTA~\cite{yuan2023robust} follows a similar but slower trajectory (35.5\% $\rightarrow$ 84.5\%). Among stable methods, PeTTA~\cite{hoang2024persistent} maintains a consistent ${\sim}35$\% but shows no improvement over time. TRIBE~\cite{Su_Xu_Jia_2024} starts competitively (33.8\% at Round~1) but degrades steadily to 44.9\% by Round~20, indicating that strong single-pass performance does not guarantee long-term stability. PeTTA+MCM achieves the best result at every round from Round~3 onward and is the only method whose error \emph{decreases} over time (33.8\% $\rightarrow$ 32.5\%), reaching a 20-round average of 32.6\%. We attribute this progressive improvement to the multi-cluster memory accumulating increasingly representative samples as corruption types recur, providing higher-quality supervision in later rounds.

\begin{table}[t]
\centering
\caption{%
\textbf{CIFAR-100-C, recurring TTA (severity~5).} 
Per-round classification error rates~(\%) over 20 successive revisits to the corruption stream.
Results use a ResNeXt-29 backbone with \textsc{RobustBench}~\cite{CroceEtAl2021_RobustBench} preprocessing. 
\textbf{Bold}~=~best; \underline{underline}~=~second best per column.}
\label{tab:recurring-cifar100}
\resizebox{\linewidth}{!}{%
\begin{tabular}{l|cccccccccccccccccccc|c}
\toprule
\textbf{Method} & 1 & 2 & 3 & 4 & 5 & 6 & 7 & 8 & 9 & 10 & 11 & 12 & 13 & 14 & 15 & 16 & 17 & 18 & 19 & 20 & Avg \\
\midrule
Source & \multicolumn{21}{c}{46.5} \\
\midrule
LAME & \multicolumn{21}{c}{40.5} \\
\midrule
CoTTA & 53.4 & 58.4 & 63.4 & 67.6 & 71.4 & 74.9 & 78.2 & 81.1 & 84.0 & 86.7 & 88.8 & 90.7 & 92.3 & 93.5 & 94.7 & 95.6 & 96.3 & 97.0 & 97.3 & 97.6 & 83.1 \\
EATA & 88.5 & 95.0 & 96.8 & 97.3 & 97.4 & 97.2 & 97.2 & 97.3 & 97.4 & 97.5 & 97.5 & 97.5 & 97.6 & 97.7 & 97.7 & 97.7 & 97.8 & 97.8 & 97.7 & 97.7 & 96.9 \\
RMT & 50.5 & 48.6 & 47.9 & 47.4 & 47.3 & 47.1 & 46.9 & 46.9 & 46.6 & 46.8 & 46.7 & 46.5 & 46.5 & 46.6 & 46.5 & 46.5 & 46.5 & 46.5 & 46.5 & 46.5 & 47.1 \\
MECTA & 44.8 & 44.3 & 44.6 & 43.1 & 44.8 & 44.2 & 44.4 & 43.8 & 43.8 & 43.9 & 44.6 & 43.8 & 44.4 & 44.6 & 43.9 & 44.2 & 43.8 & 44.4 & 44.9 & 44.2 & 44.2 \\
RoTTA & 35.5 & 35.2 & 38.5 & 41.9 & 45.3 & 49.2 & 52.0 & 55.2 & 58.1 & 61.5 & 64.6 & 67.5 & 70.7 & 73.2 & 75.4 & 77.1 & 79.2 & 81.5 & 82.8 & 84.5 & 61.4 \\
RDumb & 36.7 & 36.7 & 36.6 & 36.6 & 36.7 & 36.8 & 36.7 & 36.5 & 36.6 & 36.5 & 36.7 & 36.6 & 36.5 & 36.7 & 36.5 & 36.6 & 36.6 & 36.7 & 36.6 & 36.5 & 36.6 \\
ROID & 76.4 & 76.4 & 76.2 & 76.2 & 76.3 & 76.1 & 75.9 & 76.1 & 76.3 & 76.3 & 76.6 & 76.3 & 76.8 & 76.7 & 76.6 & 76.3 & 76.2 & 76.0 & 75.9 & 76.0 & 76.3 \\
TRIBE & \underline{33.8} & \underline{33.3} & 35.3 & \underline{34.9} & 35.3 & \underline{35.1} & 37.1 & 37.2 & 37.2 & 39.1 & 39.2 & 41.1 & 41.0 & 43.1 & 45.1 & 45.1 & 45.0 & 44.9 & 44.9 & 44.9 & 39.6 \\
PeTTA & 35.8 & 34.4 & \underline{34.7} & 35.0 & \underline{35.1} & \underline{35.1} & \underline{35.2} & \underline{35.3} & \underline{35.3} & \underline{35.3} & \underline{35.2} & \underline{35.3} & \underline{35.2} & \underline{35.2} & \underline{35.1} & \underline{35.2} & \underline{35.2} & \underline{35.2} & \underline{35.2} & \underline{35.2} & \underline{35.1} \\
\midrule
\rowcolor{cyan!10}
PeTTA + MCM & \textbf{33.8} & \textbf{33.8} & \textbf{33.0} & \textbf{33.0} & \textbf{33.1} & \textbf{33.9} & \textbf{33.9} & \textbf{33.9} & \textbf{32.7} & \textbf{32.7} & \textbf{32.7} & \textbf{32.8} & \textbf{32.7} & \textbf{32.7} & \textbf{32.7} & \textbf{32.6} & \textbf{32.6} & \textbf{32.6} & \textbf{32.6} & \textbf{32.5} & \textbf{32.6} \\
\bottomrule
\end{tabular}
}
\end{table}

\paragraph{CIFAR-10-C (Table~\ref{tab:recurring-cifar10}).}
On CIFAR-10-C, TRIBE~\cite{Su_Xu_Jia_2024} dominates with 17.5\% average error. As discussed in the main paper (Sec.~4.2), TRIBE employs a tri-net self-training architecture with balanced batch normalization that adapts directly on the current mini-batch without maintaining a memory bank; MCM, as a memory-side module, is orthogonal to this design. Among memory-based methods, PeTTA+MCM reduces the average error from 22.8\% (PeTTA alone) to 20.4\%, a 2.4-point gain. The more moderate improvement compared to CIFAR-100-C is consistent with the main paper's finding that gains scale with distributional complexity: CIFAR-10-C has only 10 classes, limiting the degree of multi-modality in the stream and thus the headroom for multi-cluster organization. Nonetheless, PeTTA+MCM is the second-best method overall and shows stable behavior across all 20 rounds, with no sign of degradation.

\begin{table}[t]
\centering
\caption{%
\textbf{CIFAR-10-C, recurring TTA (severity~5).} 
Per-round classification error rates~(\%) over 20 revisits.
All methods use a WideResNet-28 backbone from \textsc{RobustBench}~\cite{CroceEtAl2021_RobustBench} with its official preprocessing.}
\label{tab:recurring-cifar10}
\resizebox{\linewidth}{!}{%
\begin{tabular}{l|cccccccccccccccccccc|c}
\toprule
\textbf{Method} & 1 & 2 & 3 & 4 & 5 & 6 & 7 & 8 & 9 & 10 & 11 & 12 & 13 & 14 & 15 & 16 & 17 & 18 & 19 & 20 & Avg \\
\midrule
Source & \multicolumn{21}{c}{43.5} \\
\midrule
LAME & \multicolumn{21}{c}{31.1} \\
\midrule
CoTTA & 82.2 & 85.6 & 87.2 & 87.8 & 88.2 & 88.5 & 88.7 & 88.7 & 88.9 & 88.9 & 88.9 & 89.2 & 89.2 & 89.2 & 89.1 & 89.2 & 89.2 & 89.1 & 89.3 & 89.3 & 88.3 \\
EATA & 81.6 & 87.0 & 88.7 & 88.7 & 88.9 & 88.7 & 88.6 & 89.0 & 89.3 & 89.6 & 89.5 & 89.6 & 89.7 & 89.7 & 89.3 & 89.6 & 89.6 & 89.8 & 89.9 & 89.4 & 88.8 \\
RMT & 77.5 & 76.9 & 76.5 & 75.8 & 75.5 & 75.5 & 75.4 & 75.4 & 75.5 & 75.3 & 75.5 & 75.6 & 75.5 & 75.5 & 75.7 & 75.6 & 75.7 & 75.6 & 75.7 & 75.8 & 75.8 \\
MECTA & 72.2 & 82.0 & 85.2 & 86.3 & 87.0 & 87.3 & 87.3 & 87.5 & 88.1 & 88.8 & 88.9 & 88.9 & 88.6 & 89.1 & 88.7 & 88.8 & 88.5 & 88.6 & 88.3 & 88.8 & 86.9 \\
RoTTA & 24.6 & 25.5 & 29.6 & 33.6 & 38.2 & 42.8 & 46.2 & 50.6 & 52.2 & 54.1 & 56.5 & 57.5 & 59.4 & 60.2 & 61.7 & 63.0 & 64.8 & 66.1 & 68.2 & 70.3 & 51.3 \\
RDumb & 31.1 & 32.1 & 32.3 & 31.6 & 31.9 & 31.8 & 31.8 & 31.9 & 31.9 & 32.1 & 31.7 & 32.0 & 32.5 & 32.0 & 31.9 & 31.6 & 31.9 & 31.4 & 32.3 & 32.4 & 31.9 \\
ROID & 72.7 & 72.6 & 73.1 & 72.4 & 72.7 & 72.8 & 72.7 & 72.7 & 72.9 & 72.8 & 72.9 & 72.9 & 72.8 & 72.5 & 73.0 & 72.8 & 72.5 & 72.5 & 72.7 & 72.7 & 72.7 \\
TRIBE & \textbf{15.3} & \textbf{16.6} & \textbf{16.6} & \textbf{16.3} & \textbf{16.7} & \textbf{17.0} & \textbf{17.3} & \textbf{17.4} & \textbf{17.4} & \textbf{18.0} & \textbf{17.9} & \textbf{18.0} & \textbf{17.9} & \textbf{18.6} & \textbf{18.2} & \textbf{18.8} & \textbf{18.0} & \textbf{18.2} & \textbf{18.4} & \textbf{18.0} & \textbf{17.5} \\
PeTTA & 24.3 & 23.0 & 22.6 & 22.4 & 22.4 & 22.5 & 22.3 & 22.5 & 22.8 & 22.8 & 22.6 & 22.7 & 22.7 & 22.9 & 22.6 & 22.7 & 22.6 & 22.8 & 22.9 & 23.0 & 22.8 \\
\midrule
\rowcolor{cyan!10}
PeTTA+MCM & \underline{21.7} & \underline{21.0} & \underline{20.4} & \underline{19.8} & \underline{20.7} & \underline{20.1} & \underline{20.9} & \underline{20.2} & \underline{20.1} & \underline{20.4} & \underline{20.5} & \underline{20.3} & \underline{20.1} & \underline{20.0} & \underline{19.8} & \underline{20.1} & \underline{20.0} & \underline{20.3} & \underline{20.5} & \underline{20.7} & \underline{20.4} \\
\bottomrule
\end{tabular}}
\end{table}

\section{Visualization of Memory Sample Distribution}
\label{appdx:visualization}

\begin{figure}[t]
    \centering
    \includegraphics[width=1.0\linewidth]{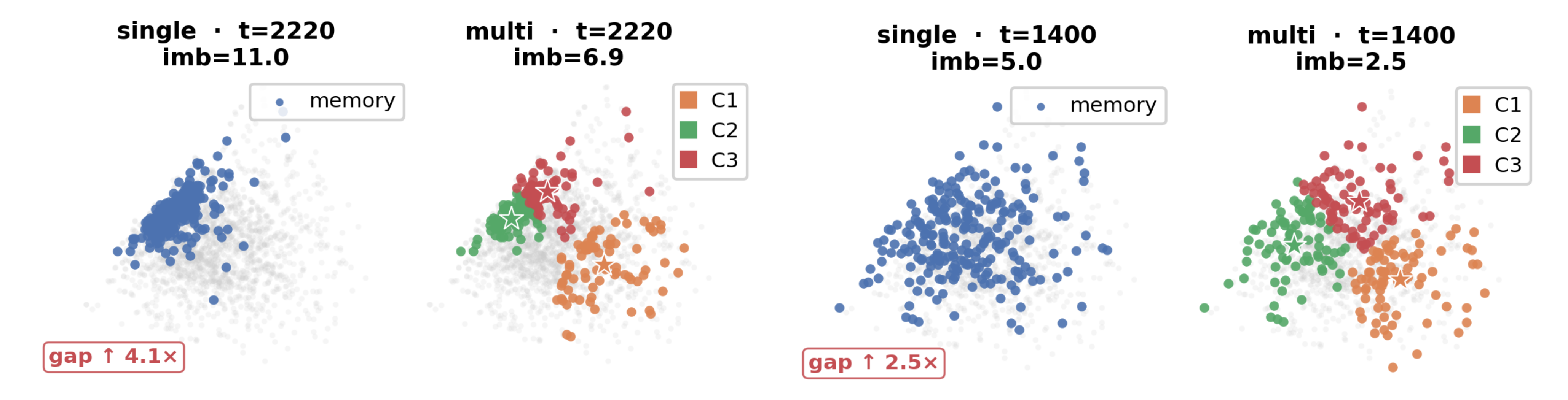}
    \caption{\textbf{Descriptor-space visualization of memory contents under SCM and MCM on CIFAR-100-C (PTTA, severity~5).} Grey points denote the incoming test stream; coloured points are stored memory samples (blue for SCM; orange/green/red for distinct MCM clusters). SCM concentrates its budget in a narrow region, yielding high imbalance ratios (imb). MCM spreads samples across distinct clusters, reducing imbalance by $4.1\!\times$ and $2.5\!\times$ at $t{=}2220$ and $t{=}1400$, respectively.}
    \label{fig:supp_mem_compare}
\end{figure}

To provide intuitive insight into how MCM organizes its memory budget, we visualize the descriptor-space distribution of stored samples at two representative time steps on CIFAR-100-C (Fig.~\ref{fig:supp_mem_compare}). We extract the channel-wise pixel-statistics descriptor (main paper, Sec.~3.2) for every sample in the memory bank and project them into two dimensions via PCA. Both snapshots ($t{=}2220$ and $t{=}1400$) correspond to transitions between consecutive corruption types, where domain diversity within the buffer is highest.

Under SCM, reservoir sampling over-represents the most recent domain, concentrating stored samples in a narrow descriptor-space region and producing high imbalance ratios ($\text{imb}{=}11.0$ and~$5.0$, respectively). MCM instead partitions the budget across dynamically formed clusters, each anchored by a centroid~($\star$), spreading samples over a substantially wider area of the descriptor manifold and reducing imbalance by $4.1\!\times$ and $2.5\!\times$. This broader coverage yields more representative mini-batches during adaptation, directly contributing to the lower error rates reported in the main paper (Tables~1--3).

\end{document}